\documentclass[conference]{IEEEtran}
\IEEEoverridecommandlockouts
\usepackage{cite}
\usepackage{amsmath,amssymb,amsfonts}
\usepackage{algorithmic}
\usepackage{graphicx}
\usepackage{textcomp}
\usepackage{xcolor}

\usepackage{colortbl}
\usepackage{algorithm}
\usepackage{algorithmic}
\usepackage{xspace}
\usepackage{multirow}
\usepackage{color}
\usepackage{subfig}
\newcommand{\stdmode}[1]{\textcolor{black!60}{#1}}
\newcommand{\vit}{Vision Transformer\xspace} 
\newcommand{\edvit}{ED-ViT\xspace} 
\usepackage{url}
\usepackage{booktabs}

\def\BibTeX{{\rm B\kern-.05em{\sc i\kern-.025em b}\kern-.08em
    T\kern-.1667em\lower.7ex\hbox{E}\kern-.125emX}}
\begin{document}

\title{Efficient Partitioning Vision Transformer on Edge Devices for Distributed Inference
\thanks{Xiang Liu and Yijun Song contribute equally to this work. Yijun Song (yijunsong.0377@gmail.com) and Linshan Jiang (linshan@nus.edu.sg) are the corresponding authors. Dr. Jialin Li is supported by the Singapore Ministry of Education Academic Research Fund Tier 1 (T1 251RES2104) and Tier 2 (MOE-T2EP20222-0016).}
}

\author{\IEEEauthorblockN{
Xiang Liu\IEEEauthorrefmark{1},
Yijun Song\IEEEauthorrefmark{2},
Xia Li\IEEEauthorrefmark{3},
Yifei Sun\IEEEauthorrefmark{4},
Huiying Lan\IEEEauthorrefmark{5},\\
Zemin Liu\IEEEauthorrefmark{4},
Linshan Jiang\IEEEauthorrefmark{6},
Jialin Li\IEEEauthorrefmark{1}}
\IEEEauthorblockA{\IEEEauthorrefmark{1}School of Computing, National University of Singapore}
\IEEEauthorblockA{\IEEEauthorrefmark{2}Information and Artificial Intelligence Institute, Zhejiang University of Finance \& Economics Dongfang College}
\IEEEauthorblockA{\IEEEauthorrefmark{3}Department of Computer Science, ETH Zurich}
\IEEEauthorblockA{\IEEEauthorrefmark{4}College of Computer Science and Technology, Zhejiang University} \IEEEauthorblockA{\IEEEauthorrefmark{5}Lumia Ltd.}\IEEEauthorblockA{\IEEEauthorrefmark{6}Institute of Data Science, National University of Singapore}
}

\maketitle

\begin{abstract}
Deep learning models are increasingly utilized on resource-constrained edge devices for real-time data analytics. Recently, Vision Transformer and their variants have shown exceptional performance in various computer vision tasks. However, their substantial computational requirements and low inference latency create significant challenges for deploying such models on resource-constrained edge devices. To address this issue, we propose a novel framework, ED-ViT, which is designed to efficiently split and execute complex Vision Transformers across multiple edge devices. Our approach involves partitioning Vision Transformer models into several sub-models, while each dedicated to handling a specific subset of data classes. To further reduce computational overhead and inference latency, we introduce a class-wise pruning technique that decreases the size of each sub-model. Through extensive experiments conducted on five datasets using three model architectures and actual implementation on edge devices, we demonstrate that our method significantly cuts down inference latency on edge devices and achieves a reduction in model size by up to 28.9 times and 34.1 times, respectively, while maintaining test accuracy comparable to the original Vision Transformer. Additionally, we compare ED-ViT with two state-of-the-art methods that deploy CNN and SNN models on edge devices, evaluating metrics such as accuracy, inference time, and overall model size. Our comprehensive evaluation underscores the effectiveness of the proposed ED-ViT framework. 

\end{abstract}

\begin{IEEEkeywords}
Distributed Inference, Edge Computing, Model Splitting, Vision Transformer
\end{IEEEkeywords}


\section{Introduction}
In recent years, deep learning models have been increasingly deployed on resource-constrained edge devices to meet the growing demand for real-time data analytics in industrial systems~\cite{chen2021split,chen2023nnfacet,Yu2024ECSNNSD} and have demonstrated remarkable capabilities in various applications such as video image analysis and speech recognition. Convolutional Neural Networks (CNNs)~\cite{krizhevsky2012imagenet} like VGGNet~\cite{simonyan2014very} and ResNet~\cite{He2015DeepRL}, as well as Spike Neural Networks (SNNs)~\cite{Deng2021OptimalCO}, have achieved satisfactory performance in many edge computing scenarios. As the field progresses, researchers are exploring the deployment of more complex structured models on edge devices to further improve performance. Transformer architecture~\cite{vaswani2017attention}, which has revolutionized natural language processing (NLP) tasks, has inspired similar advancements in computer vision. Vision Transformer (ViT) models~\cite{alexey2020image} and their variants have shown outstanding results across various computer vision tasks, including image classification~\cite{wu2020lite,touvron2021training}, object detection~\cite{carion2020end,dai2021up,yang2021uncertainty}, semantic segmentation~\cite{song2020vr,wang2021end} and action recognition~\cite{girdhar2019video,plizzari2021spatial} and audio spectrogram recognition~\cite{gong2022ssast}. The success of ViTs has sparked interest in leveraging their capabilities for edge computing applications.

However, the rapid advancement in machine learning technologies has increased the demand for computational resources and memory, given the complexity of these model configurations. Achieving higher accuracy with ViT requires substantial computational power and memory, which poses challenges for deployment on edge devices. For instance, ViT-Base~\cite{alexey2020image} consists of over 86.7 million parameters and requires approximately 330MB of memory. Researchers now face the dilemma of deploying such complex models while dealing with resource-constrained resources.

Previous studies aiming to reduce the deployment overhead primarily focus on compressing \vit models. These approaches can be classified into three major categories: (1) architecture and hierarchy restructuring~\cite{Pan2021ScalableVT,Graham2021LeViTAV}, (2) encoder block enhancements~\cite{Zhang2021MultiScaleVL,Yu2021MetaFormerIA,Yang2021LiteVT,Tu2022MaxViTMV,Yao2022DualVT,Pan2023SlideTransformerHV}, and (3) integrated approaches~\cite{Yuan2021IncorporatingCD,Dai2021CoAtNetMC}. However, these methods often suffer from either poor inference accuracy or high inference latency as they attempt to compress a large model to fit into a memory-constrained edge device.

\begin{figure*}[!ht]
  \centering
  \includegraphics[width=\textwidth]{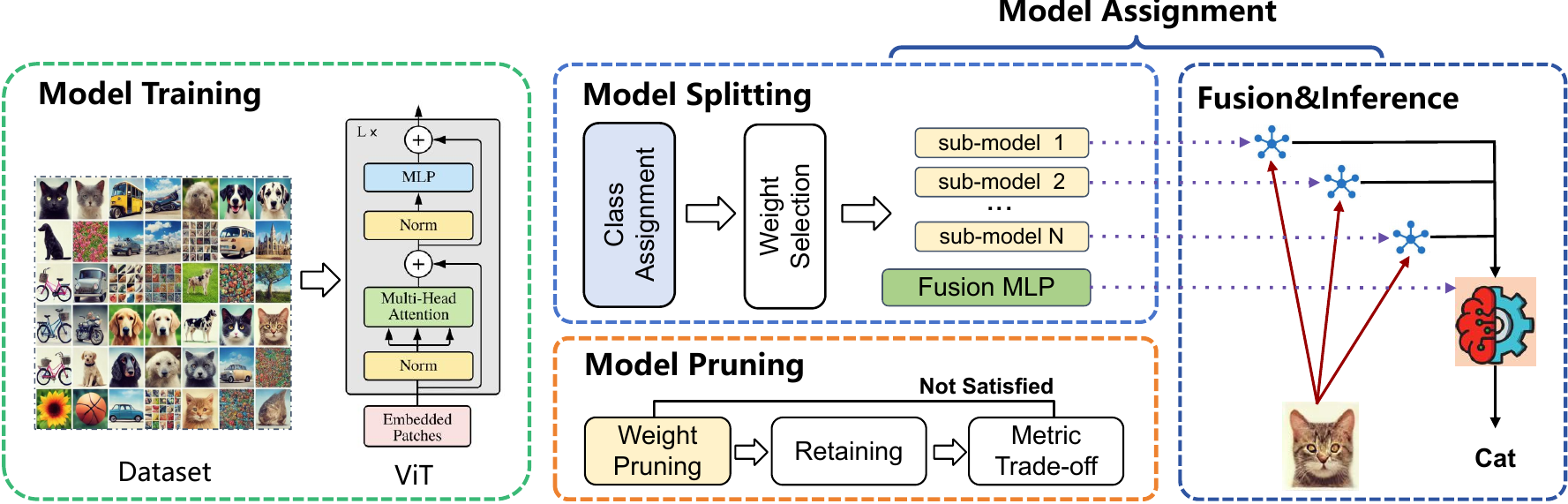}
  \caption{The overview of \edvit, including four steps: Model Splitting, Model Pruning, Model Assignment and Model Fusion.}
  \label{fig:overview}
\end{figure*}

To develop a solution that mitigates accuracy drop and enables efficient deployment of the \vit on resource-constraint edge devices, we aim to utilize the collaboration of multiple edge devices and propose a \vit splitting framework named \textbf{E}dge \textbf{D}evice \textbf{Vi}sion \textbf{T}ransformer, shorten as \textbf{ED-ViT}. As illustrated in Fig.~\ref{fig:overview}, \edvit first partitions the original \vit into several smaller sub-models, inspired by the concept of split learning (SL). However, unlike traditional SL, which does not consider edge device constraints, each of these small sub-models is responsible for detecting a specific subset of the classes and is deployed on resource-constrained edge devices. \edvit then employs model pruning techniques to further alleviate the computational load and processing requirement for each sub-model. To optimize model assignment, we design a greedy assignment algorithm that takes into account both the model computational resources and memory resources. Besides integrating the previous steps, \edvit uses a multilayer perceptron (MLP) model to fuse the results from all the sub-models. We conduct experiments across five datasets to evaluate the effectiveness of \edvit framework on edge devices, particularly in low-power video analytics. The results, measured across three key metrics—accuracy, inference latency, and model size—consistently demonstrate the significant benefits of \edvit. Additionally, we compare \edvit with methods that split CNN and SNN, highlighting the great potential of deploying \vit on resource-constrained edge devices to achieve high accuracy while maintaining small model sizes and low inference latency. Our main contributions are summarized as follows:

\begin{itemize} 
\item This is the first study focusing on combining pruning and splitting, deploying \vit onto edge devices. We propose a framework that leverages the capabilities of \vit, allowing for the collaboration of multiple edge devices to achieve distributed inference in practical applications. 


\item We introduce \edvit to address the \vit splitting problem by decomposing the complex original model into sub-models, and applying pruning techniques to reduce the size of each sub-model. Using a combined greedy method for model assignment, \edvit effectively addresses the formulated problem by reducing model sizes, minimizing inference latency, and maintaining high accuracy, achieving a trade-off across these three metrics.

\item We conduct extensive experiments with three computer vision datasets and two audio recognition datasets across three ViT structures. Besides, we implement our \edvit on Raspberry Pi 4Bs, demonstrating that our framework significantly reduces inference latency on edge devices and decreases overall memory usage with negligible accuracy loss in various applications.
\end{itemize}

\section{Related Works}
\label{sec:related}

\subsection{General \vit Compression}
Deploying \vit models in resource-constrained environments poses significant challenges due to their intensive computational and memory demands. These approaches address \vit resource limitations via pruning, encompassing both local and global strategies as follows.

\begin{table*}[t!]
\centering
\caption{Performance characteristics for standard \vit models with their default parametrization at resolution 224 $\times$ 224: ViT-Small, ViT-Base and ViT-Large, all with patch size of 16 $\times$ 16. All the experiment results are obtained on Raspberry Pi-4B devices.
}
{\normalsize
\begin{tabular}{c|ccc|cccc}
\hline
\multirow{2}{*}{\textbf{Model}}  & \multirow{2}{*}{\textbf{Depth}} & \multirow{2}{*}{\textbf{Width}}  & \multirow{2}{*}{\textbf{Heads}}  & \textbf{Params} & \textbf{Flops} & \textbf{Latency} & \textbf{Mem Size} \\
&  &  &  & \ ($\times 10^6$) & ($\times 10^9$)  & (ms)  & (MB) \\
\hline
ViT-Small  & 12   & 384  & 6  & 22.1  & 4.25 & 9628 & 83\\
ViT-Base   & 12   & 768  & 12  & 86.6 & 16.86 & 36940 & 327\\
ViT-Large  & 24   & 1024 & 16  & 304.4  & 59.69& 118828 & 1157\\
\hline
\end{tabular}
}
\label{table:vit}
\end{table*}

\textbf{Local pruning techniques} focus on removing redundant components within specific layers of the model. For instance, PVT~\cite{Wang2021PyramidVT} and its successor PVTv2~\cite{Wang2021PVTVI} introduce a pyramid hierarchical structure to transformer backbones, achieving high accuracy with reduced computation. \cite{zhu2021vision} applies sparsity regularization during training and subsequently prunes the dimensions of linear projections, targeting less significant parameters. \cite{xia2022structured} prunes multi-head self-attention (MHSA) and feed-forward networks (FFN), which are often redundant components.  \cite{liang2022not,Liu2023RevisitingTP} propose network pruning to eliminate complexity and model sizes by reducing tokens. Other noteworthy contributions include DToP~\cite{tang2023dynamic}, which enables early token exits for semantic segmentation tasks. Conversely, \textbf{global pruning techniques} adopt a comprehensive perspective by evaluating and pruning the overall significance of neurons or layers across the entire network. SAViT~\cite{Zheng2022SAViTSV} purposes structure-aware \vit pruning via collaborative optimization. For instance, CP-ViT~\cite{song2022cp} systematically assesses the importance of head and attention layers for the purpose of pruning, while Evo-ViT~\cite{xu2022evo} identifies and preserves significant tokens, thereby discarding those of lesser importance. Moreover, the Skip-attention approach~\cite{venkataramanan2023skip} facilitates the omission of entire self-attention layers, thereby exemplifying a global pruning methodology. X-pruner~\cite{yu2023x} employs explainability-aware masks to inform its pruning decisions, thereby advancing a more informed global pruning strategy. In addition, UP-ViT~\cite{yu2023unified} introduces a unified pruning framework that leverages KL divergence to guide the decision-making process for pruning, while LORS~\cite{li2024lors} optimizes parameter usage by sharing the majority of parameters across stacked modules, thereby necessitating fewer unique parameters. 

Among existing pruning methods, UP-ViT~\cite{yu2023unified} has the closest resemblance to our approach. However, it is important to note that these techniques cannot be directly applied to edge devices: they often suffer from poor performance when the pruning ratio is high or incur high computation overhead when the pruning ratio is low, making them unsuitable for resource-constrained edge environments. In contrast, our work introduces a class-based global structured pruning method that addresses these limitations. Our approach is orthogonal to most previous methods and does not involve trainable parameters, contributing to more stable performance.

\subsection{\vit on Edge Devices}
There are several methods focused on deploying \vit on-edge devices, which can be classified into three major categories.

\textbf{Architecture and Hierarchy Restructuring: }HVT~\cite{Pan2021ScalableVT} compresses sequential resolutions using hierarchical pooling, reducing computational cost and enhancing model scalability. LeViT~\cite{Graham2021LeViTAV} is a hybrid model that combines the strengths of CNNs and transformers. For image classification tasks, it utilizes the hierarchical structure of LeNet~\cite{krizhevsky2012imagenet} to optimize the balance between accuracy and efficiency, and uses average pooling in the feature map stage. MobileViTv3~\cite{wadekar2022mobilevitv3} propose changes to the fusion block, which addresses the scaling and simplifies the learning tasks. 

\textbf{Encoder Block Enhancements: }ViL~\cite{Zhang2021MultiScaleVL} introduces a multiscale vision longformer that lessens computational and memory complexity when encoding high-resolution images. Poolformer~\cite{Yu2021MetaFormerIA} deliberately replaces the attention module in transformers with a simple pooling layer. LiteViT~\cite{Yang2021LiteVT} introduces a compact transformer backbone with two new lightweight self-attention modules (self-attention and recursive atrous self-attention) to mitigate performance loss. Dual-ViT~\cite{Yao2022DualVT} reduces feature map resolution, consisting of two dual-block and two merge-block stages. MaxViT~\cite{Tu2022MaxViTMV} divides attention into local and global components and decomposes it into a sparse form with window and grid attention. Slide-Transformer~\cite{Pan2023SlideTransformerHV} proposes a slide attention module to address the problem that computational complexity increases quadratically with the attention modules, while EdgeViT~\cite{pan2022edgevits} enables attention-based vision models to compete with the best light-weight CNNs when considering the tradeoff between accuracy and on-device efficiency. 

\textbf{Integrated Approaches: }Some methods integrate both of the above approaches. CeiT~\cite{Yuan2021IncorporatingCD} combines Transformer and CNN strengths to overcome the shortcomings of each, incorporating an image-to-tokens module, locally-enhanced feedforward layers, and layer-wise class token attention. CoAtNet~\cite{Dai2021CoAtNetMC} combines depth-wise convolutions and simplifies traditional self-attention by relative attention, enhancing efficiency by stacking convolutions and attention layers. DeViT~\cite{xu2023devit} also decomposes \vit 
 for collaborative inference. However, DeViT trains a ViT-Large for each sub-model even when splitting ViT-Small and employs model distillation to enhance accuracy, which introduces significant training overhead. In addition, the smallest model size that DeViT provides is larger than 90MB.

However, they never consider linking pruning with specific classes, which limits their methods when both high performance and low memory usage are required.

\subsection{Split Learning}
Current works that combine \vit and split learning primarily focus on federated learning, addressing data privacy and efficient collaboration in multi-client environments~\cite{Oh2022DifferentiallyPC, Almalik2023FeSViBSFS,Oh2024PrivacyPreservingSL}, where the inner structure of a large model is split across smaller devices and later fused~\cite{Su2022AdaptiveST}. However, these approaches do not target the deployment of \vit on edge devices.

Traditional machine learning model splitting generally involves partitioning a large model into multiple smaller sub-models that can be executed collaboratively on resource-constrained devices, providing a promising technique for deploying models on edge devices. Splitnet~\cite{Kim2017SplitNetLT} clusters classes into groups, partitioning a deep neural network into tree-structed sub-networks. \cite{Bakhtiarnia2022DynamicSC} dynamically partitions models based on the communication channel's state. Nnfacet~\cite{chen2021split,chen2023nnfacet} splits large CNNs into lightweight class-specific sub-models to accommodate device memory and energy constraints, with the sub-models being fused later. \cite{Yu2024ECSNNSD} follows a similar approach to split deep SNNs across edge devices. Distredge~\cite{Hou2022DistrEdgeSU} uses deep reinforcement learning to compute the optimal partition for CNN models.

To the best of our knowledge, our work presents the first exploration of \vit model partitioning for edge deployment, marking a significant contribution to this field. Drawing inspiration from previous studies~\cite{chen2021split,chen2023nnfacet,Yu2024ECSNNSD}, our framework, \edvit, introduces an innovative approach to decompose a multi-class ViT model into several class-specific sub-models, each performing a subset of classification. Unlike relying on channel-wise pruning,  \edvit employs advanced pruning techniques specifically designed for the unique architecture of Vision Transformers.

\section{Problem Formulation}
\label{sec:pro}

The structures of three representative \vit, ViT-Small, ViT-Base, and ViT-Large, are presented in Table~\ref{table:vit}. The number of operations is commonly used to estimate computational energy consumption at the hardware level. In \vit models, almost all floating-point operations (FLOPs) are multiply-accumulate (MAC) operations. 

For Patch Embedding, FFN, and MLP Head, their operation counts are easy to infer as they follow a fully connected (FC) structure, where the MAC count is given by $(2FC_{in} + 1) \times FC_{out}$, where $FC_{in}$ and $FC_{out}$ represent the input and output features, respectively. For MHSA, assuming the number of patches is $p$, the dimension of each patch is $d_p$, the embedding dimension is $d$, and the number of attention heads is $h$, the MAC for the linear projections to generate the $Q$, $K$, and $V$ matrices is $3 \times p \times d^2 / h$. The MAC for $QK^T$ is $p^2 \times d / h$, and the MAC for the softmax operation and multiplication with $V$ is $p^2 \times d / h$. For $h$ attention heads, the total MAC is $h \times (3 \times p \times d^2 / h + 2 \times p^2 \times d / h) = 3 \times p \times d^2 + 2 \times p^2 \times d$. Based on this analysis, energy consumption can be estimated as being proportional to FLOPs, given that the pruned model follows the same structure.

Thus, we formulate the problem as follows. We assume that we have $L$ inference samples in total to be processed, and the set of $N$ edge devices is represented as $D$. The available memory and energy (FLOPs of an edge device)~\cite{green_machines} for each edge device $D_i$ are denoted as $M_i$ and $E_i$, respectively. The FLOPs (energy consumption) for each inference sample for the sub-model $Model_j$ from the set of sub-models $\{Models\}$ is represented as $e_j$, calculated based on the previous energy consumption estimation. To formulate the problem of \vit partitioning and edge-device-based deployment, we define the objective function as $max_{\{Model_j\}} min_{D_i \in D} \{E_i-Le_j\}$, aiming to minimize the maximal inference latency, as inference latency is closely related to the computational power of edge devices. Additionally, the accuracy $a_{fus}$ of the fused results from all $N$ inference samples must be greater than or equal to the required inference accuracy $A_{re}$; the total memory sizes of all sub-models should not exceed the memory budget $bu$.

The optimization problem can be formally formulated as follows, where $x_{ie}$ is a binary decision variable: $1$ indicates that the sub-model deployed on edge device $D_i$ is responsible for class $e$, and $0$ otherwise. Each sub-model learns a specific subset of the classes in $C$. Furthermore, the memory consumption of sub-model $j$, denoted as $size$(Model$_j$), represented as $m_j$, must be smaller than the available memory size of the deployed edge device:

\begin{equation}
\label{eq:problem}
\left\{
\begin{aligned}
    &argmax_{\{Model_j\}} min_{D_i \in D} \{E_i-Le_j\}\\
    & \text{s.t.} \quad 
     L  e_j \leq E_i, \quad 
     \text{Model}_j \quad\text{deploys on} \quad D_i \\
    & \quad \quad   m_j \leq M_i, \quad  
    \\
    & \quad  \quad  a_{fus} \geq A_{re}, \\
    & \quad \quad \sum_{j} m_j \leq bu,  \\
    & \quad  \quad  \sum_{i=1}^{|D|} x_{ie} = 1, \quad \forall e \in C, \forall i \in D
\end{aligned}
\right.
\end{equation}

\begin{figure*}[ht!]
\centering
  \includegraphics[width=\textwidth]{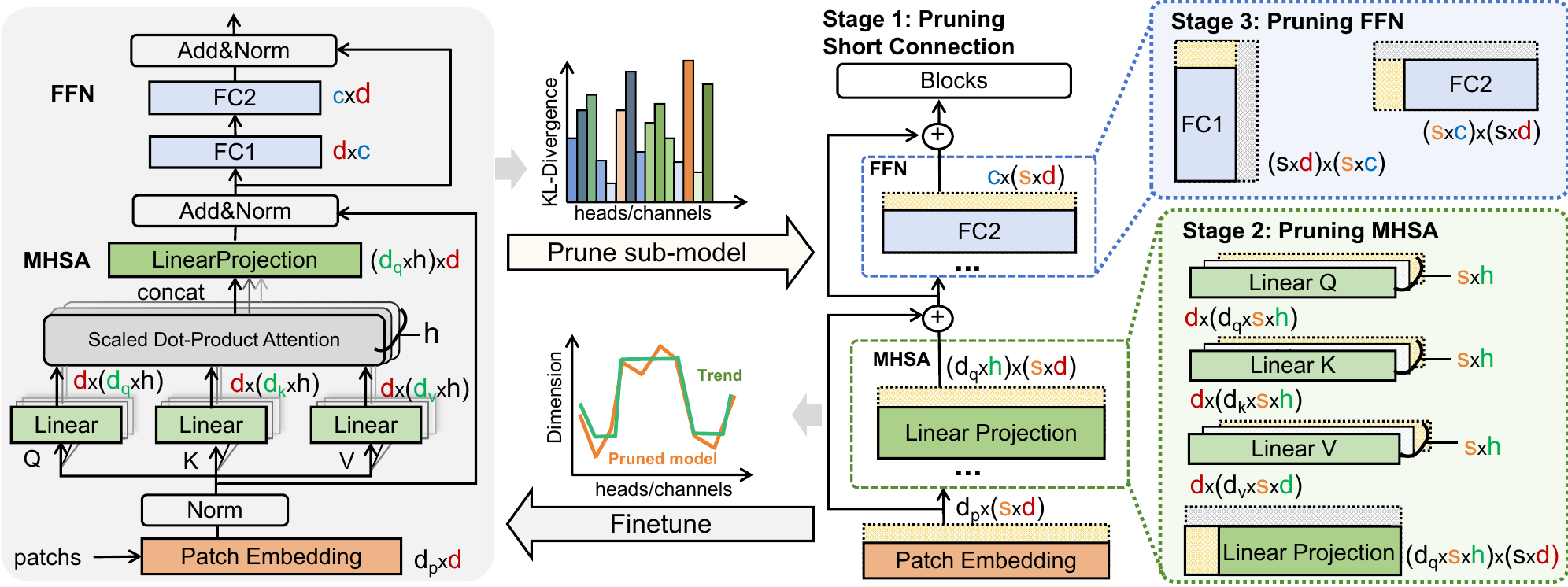}
\caption{
Structured pruning of a \vit block. Left: illustration of prunable components in a ViT block. Right: corresponding sequential pruning process. Our approach targets three key components: (1) channels in residual connections (red, denoted as $d$), (2) the number of heads in the MHSA module (green, denoted as $h$), and (3) hidden layer channels in the FFN (blue, denoted as $c$). The pruning process occurs in three stages: residual connection channels, MHSA heads, and FFN hidden dimensions. Yellow regions indicate parameters being pruned in the current stage, while gray regions represent previously pruned parameters.
}
\label{fig:vit_pruning}
\end{figure*}

\section{Methodology}
\label{sec:method}
This section describes the design of the \edvit framework proposed to solve the optimization problem outlined in \eqref{eq:problem}.  We first explain the main workflow of \edvit and then provide detailed descriptions of the four key steps involved.

\subsection{Design Overview}
As illustrated in Fig.~\ref{fig:overview}, \edvit leverages the unique characteristics of \vit and the collaboration of multiple edge devices. The framework involves $N$ concurrent edge devices for distributed inference alongside a lightweight MLP aggregation to derive the final classification results. Initially, the original \vit is trained on the entire dataset to achieve high test accuracy for the classification task. The \edvit framework is composed of four main components: model splitting, pruning, assignment, and fusion. During model splitting, the \vit model is divided into sub-models, each responsible for a subset of classes. To reduce computation overhead, these sub-models are further pruned using model pruning techniques. Subsequently, the sub-models are assigned to the appropriate edge devices, taking the optimization problem into consideration. Finally, the aggregation device fuses the outputs from the edge devices to produce the final inference results. The specific details of each component are provided below.

\begin{algorithm}
    \caption{Model Splitting in \edvit}
    \label{alg:modelsplitting}
    \begin{flushleft}
    \textbf{Input}: The number of edge devices $N$; memory budget $bu$; initial pruning head number $hp=hp_1, hp_2, ..., hp_N$ for all sub-models, remaining available memory size $M_i$ and remaining computational resource $E_i$ for device $i$; training dataset $\mathbf{(X, y)}$ \\
\textbf{Parameter}: the classes set $C$; trained original Model$_0$\\
\textbf{Output}: class-specific sub-models \{Model$_1$,..., Model$_N$\} and a fusion model $MLP$
\end{flushleft}
    \begin{algorithmic}[1]
        \STATE Let $flag_r=True$, $D$=\{device $1$, ..., device $N$\}.
        \STATE Let $E=\{ E_1, ..., E_N \}$, $M=\{ M_1, ..., M_N \}$.
        \REPEAT 
        \STATE Let $C=\{C_1,C_2, ..., C_N\}, s.t. |C|=\sum_{i=1}^N|C_i|$.
        \STATE $C_i$ is determined randomly.
        \UNTIL{$\forall C_a, C_b \in C, \big| |C_a| - |C_b| \big| \leq1$.}

        \WHILE{$flag_r$ is $True$}
            \FOR{$i$ in $N$}
                \STATE Model$_i=prune(\text{Model}_0, \mathbf{X, y}, C_i,hp_i)$.
            \ENDFOR
            \STATE $MA=\phi$
            \IF{$\sum_{j} m_i \leq bu$}
                 \STATE $MA=greedySearchAssign(E, M, \text{Model}_i, D)$.
            \ENDIF{}

            \IF { $MA \neq \phi$ } 
                \STATE $flag_r=False$.
            \ELSE
                \STATE $hp_j = hp_j+1$ where Model$_j$ has the biggest memory size.
            \ENDIF
        \ENDWHILE
        \STATE $concat_{out}$ = $concat$(Model$_1$($\mathbf{X}$),..., Model$_N$($\mathbf{X}$)).
        \STATE $MLP$ = $train$($concat_{out}$, $\mathbf{y}$).
        \STATE \textbf{return} $MLP$, Model$_1$, ..., Model$_N$.
    \end{algorithmic}
   
\end{algorithm}

\begin{algorithm}[tb]
    \caption{Model Pruning in \edvit}
    \label{alg:modelpruning}
        \begin{flushleft}
    \textbf{Input}: pruning head number $hp_i$; assigned classes subset $C_i$ \\
    \textbf{Parameter}: the raw original Model$_0$; training dataset $(\mathbf{X, y})$ \\
    \textbf{Output}: pruned Model$_i$
        \end{flushleft}
    \begin{algorithmic}[1] 
        \STATE $\mathbf{X_i,y_i}$ = $resample(\mathbf{X, y}, C_i)$.
        \STATE Model$_i$ = $PruneShortConnection($Model$_0,hp_i)$
        \STATE Model$_i$ = $PruneMHSA($Model$_i,hp_i)$
        \STATE Model$_i$ = $PruneFFN($Model$_i,hp_i)$
        \STATE Model$_i$ = $retrain$(Model$_i$, $X_i$, $y_i$).
        \STATE \textbf{return} Model$_i$.
    \end{algorithmic}
\end{algorithm}

\subsection{Model Splitting}
In the original \vit, different heads contribute to learning and inferring from the samples. However, for certain classes, maintaining all the connections between the heads can be redundant. As a result, \edvit prunes these connections and reconstructs the heads, with more retained heads leading to more parameters and connections being preserved. As illustrated in Algorithm~\ref{alg:modelsplitting}, each \vit sub-model undergoes pruning based on a head number threshold and its associated categories, following a relatively equitable workload distribution. Subsequently, a greedy search mechanism is used to identify the most suitable edge device model assignment plan for deploying a particular sub-model, considering both energy and memory constraints. If the total memory size exceeds the budget or no suitable plan is found, an iterative approach is applied to adjust the number of heads for the sub-model with the biggest memory size to be pruned, repeating the allocation process until all sub-models are successfully assigned to edge devices. The pruning and the greedy assignment methods are shown as Algorithm~\ref{alg:modelpruning} and Algorithm~\ref{alg:modelassignment2}, located in Section~\ref{sec:modelpru} and ~\ref{sec:modelass}, respectively.

\subsection{Model Pruning}
\label{sec:modelpru}

\noindent We believe that reducing the computational burden of \vit will significantly contribute to lowering inference latency in distributed edge device settings. We focus on the original ViT architecture~\cite{alexey2020image}, chosen for its simplicity and well-defined design space, focusing on redistributing the dimensionality across different blocks to achieve a more balanced tradeoff between computational efficiency and accuracy, as shown in Fig.~\ref{fig:vit_pruning}.  

\textbf{Analysis of Prunable Parameters: } The main prunable components in a ViT block, as illustrated in Fig.~\ref{fig:vit_pruning}.

\begin{itemize}
    \item Residual Connection Channels (Red, $d$): The channels across the shortcut connections within the transformer blocks.
    \item Heads in MHSA (Green, $h$): The dimensions of the query, key, value projections ($d_q$, $d_k$, $d_v$)\footnote{$d_q = d_k = d_v = d/h$}.
    \item Feed-Forward Network (FFN) Hidden Dimensions (Blue, $c$): The dimension $c$ of the hidden layer used for expanding and reducing.
\end{itemize}

\textbf{Pruning Process: } As illustrated in Fig.~\ref{fig:vit_pruning}, The pruning process is carried out in stages, with each stage focusing on one of the prunable components. We compute the KL-Divergence between the output distributions of the original model and the pruned model to evaluate the importance of each component, as follows:
\[
D_{\text{KL}}(P \parallel Q) = \sum_i P(i) \log \frac{P(i)}{Q(i)}
\]
where $P(i)$ represents the output distribution of the original model, and $Q(i)$ represents the distribution after pruning. 

We focus on pruning the channels of the residual connections (as shown in red) in the first stage. Using KL-Divergence, we identify and prune the channels that contribute the least, reducing the dimensionality from $d$ to $s \times d$, and the pruning factor $s$ controls the degree of reduction in the parameters. We use $j$-th sub-model as an example: we set $s=(h-hp_j)/h$, effectively controlling the extent of the pruning and parameter reduction. This helps to streamline the flow of information between layers without significantly affecting model performance. Then, instead of directly removing entire heads in the MHSA module, we prune the least important dimensions within the query, key, and value projections ($d_q$, $d_k$, and $d_v$) across multiple heads. This process effectively reduces the total number of heads to $s \times h$, without entirely discarding any head, thus maintaining a balanced representation of the attention mechanism while reducing its complexity. The dimensionality of the projections is scaled accordingly to reflect the merging and pruning process, ensuring that the model retains its ability to capture token interactions. The final stage involves pruning the hidden dimension $c$ in the FFN, as shown in blue. By calculating KL-Divergence, we identify the least important neurons and reduce the hidden dimension from $c$ to $s \times c$. Following each pruning stage, the model is fine-tuned to recover any performance loss that may result from the parameter reduction. This ensures that the pruned model achieves a similar level of accuracy as the original model while requiring fewer computational resources. 

In conclusion, the pruning process is outlined in Algorithm~\ref{alg:modelpruning}. An additional advantage of \edvit is that, even after pruning, the sub-models still retain the structure of \vit. This gives our method the potential to be combined with other horizontal pruning techniques for ViT and its variants and leverage the inherent features of \vit models to generalize well into downstream tasks.

\subsection{Model Assignment}
\label{sec:modelass} 

\begin{algorithm}[tb]
    \caption{Model Assignment in \edvit}
    \label{alg:modelassignment2}
        \begin{flushleft}
    \textbf{Input}:  remaining available memory size set $M$, remaining computational resource set $E$, the edge device set $D$, the sub-model set. \\
    \textbf{Output}: Model assignments $MA$\\
        \end{flushleft}
    \begin{algorithmic}[1] 
        \STATE $\{Models\} \leftarrow sort(\{Models\})$ (Sort the sub-models based on the computation overhead from the highest to lowest).
        
        \FOR{$i$ in $N$}
                \STATE $j \leftarrow argmax_{k \in D }E_k$.
                \IF { $M_j >= size(\text{Model}_i)$ and $E_j >= computing($\text{Model}$_i) $} 
                    \STATE $E_j \leftarrow E_j - computing($\text{Model}$_i)$.
                    \STATE $M_j \leftarrow M_j - size($\text{Model}$_i)$.
                \ELSE
                    \STATE $D \leftarrow D-j $.
                    \IF{ $D = \phi$}
                        \STATE \textbf{return} $\phi$ 
                    \ENDIF
            \ENDIF
        \ENDFOR
        \STATE \textbf{return} $MA$
    \end{algorithmic}
\end{algorithm}

\noindent To address the optimization problem expressed in \eqref{eq:problem}, we propose a greedy search algorithm for assigning \vit sub-models to edge devices. As shown in Algorithm\ref{alg:modelassignment2}, the sub-models are first sorted based on their energy consumption. \edvit assigns the most computation-intensive sub-model first based on their model sizes, which is proportional to the computation overhead as in Section~\ref{sec:pro}. The algorithm then iteratively assigns the remaining sub-models to maximize the system’s available energy. Initially, the device with the highest computational power is selected. If the remaining memory and energy can accommodate the sub-model, we update the device’s available memory and energy. Otherwise, if the sub-model exceeds the device’s memory capacity, the memory-exhausted device is removed from the set. If no devices remain, it indicates that the current pruning results prevent deployment of all sub-models. In this case, the algorithm terminates, and the \edvit framework re-prunes the sub-models based on a new head pruning parameter, as described in Algorithm~\ref{alg:modelsplitting}. Finally, the algorithm outputs the model assignment plan $MA$, representing the mapping of sub-models to edge devices.

As described in Section~\ref{sec:pro}, the problem of \vit sub-model partitioning and assignment can be formulated as a 0-1 knapsack problem, where each edge device has varying available memory and energy. Each sub-model is responsible for a specific set of classes, and multiple sub-models can be deployed on a single device. We perform a collaborative optimization of partitioning the \vit model into multiple sub-models, as shown in Algorithm~\ref{alg:modelsplitting}, and deploying these sub-models across edge devices using a greedy search assignment mechanism , as shown in Algorithm~\ref{alg:modelassignment2}. This approach provides a relatively optimal solution to the formulated problem. Our extensive experiments demonstrate the effectiveness of our framework design and algorithms.

\subsection{Model Fusion}
In the result fusion phase, each sub-model on the edge devices processes inputs and extracts corresponding features. The aggregation edge device aggregates the generated features through concatenation and feeds them into an MLP to produce the final prediction. Notably, the MLP for result fusion requires training only once after all sub-models have been trained. 

In our paper, we utilize a tower-structured MLP to process the concatenated tensors received from the various edge devices. Specifically, each transmitted tensor from a device is integrated using a $N \times d \times s $ $\rightarrow$ $\lambda \times N \times d \times s$  $\rightarrow$ $numcls$ MLP structure, where $\lambda$ is the shrinking hyperparameter and the default value is 0.5, $numcls$ is the number of classes. By utilizing a compact MLP model, we effectively fuse the distributed inference results from the sub-models while consuming only a minimal amount of computational resources.

\section{Experiments}
\label{sec:experiments}

\subsection{Experiments settings}
\label{sec:experimentsettings}

\textbf{Datasets. }Considering the versatile applicability of the framework across various scenarios, we select three computer vision datasets (i.e., CIFAR-10~\cite{Krizhevsky2009LearningML}, MNIST~\cite{LeCun1998GradientbasedLA}, and Caltech256~\cite{griffin2007caltech}) and two audio recognition datasets (i.e., GTZAN~\cite{tzanetakis2002musical} and Speech Command~\cite{warden2018speech}) to construct the classification tasks for our experiments. For all the computer vision datasets, we resize the sample to $224\times224\times3$ to support various datasets and downstream tasks via a similar data structure without loss of generality; for the audio recognition datasets, we resize the sample to $224\times224\times1$ with the same aim. 

\begin{figure}[t!]
\centering
\includegraphics[width=0.9\linewidth]{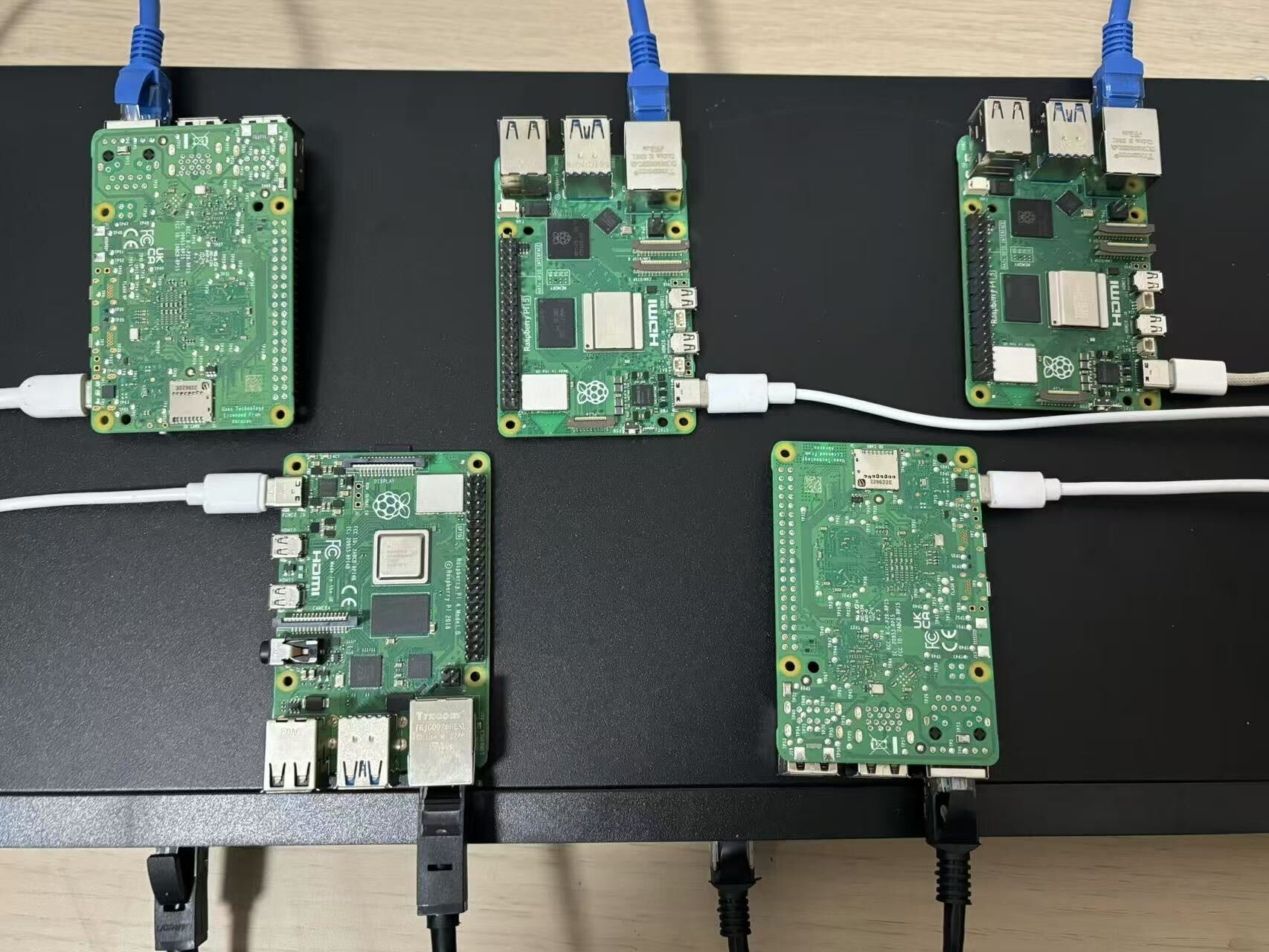}
\caption{Our 5-device example experimental prototype utilizes a switch and Raspberry Pi 4B devices, with one dedicated to the fusion model and the other four allocated to sub-models.}
\label{fig:real}
\end{figure}

\noindent\textbf{Implementation Details: } All models are implemented using Pytorch~\cite{paszke2019pytorch}. During the training process, we use the Adam optimizer~\cite{diederik2014adam} with a decaying learning rate initialized to 1$e$-4, and we set the batch size to 256. For the computer vision task, the original \vit model is pre-trained on the ImageNet dataset~\cite{deng2009imagenet}, followed by fine-tuning the task-specific data for 10 epochs. For the audio recognition task, \vit is pre-trained on the AudioSet dataset~\cite{gemmeke2017audio} and then fine-tuned on the task data for about 20 epochs. All the experimental results are averaged over five trial runs. We use the server with 8$\times$NVIDIA A100 GPUs to generate sub-models and the fusion model. Each inference trial is conducted on 1 Raspberry Pi-4B for fusion and 1 to 10 Raspberry Pi-4B devices for sub-models, which serve as the edge devices for evaluating the execution time of processing a single sample on a specific sub-model. The edge devices are all connected with a gigabyte switch S1720-52GWR-PWR-4P as shown in Fig.~\ref{fig:real}. For bandwidth control, we use the traffic control tool tc~\cite{TCFoundation2022}, which is able to limit the bandwidth under the setting value. The maximum bandwidth between devices is capped at 2 Mbps to simulate real-world scenarios. 

\begin{figure}[t!]
\centering
\includegraphics[width=\linewidth]{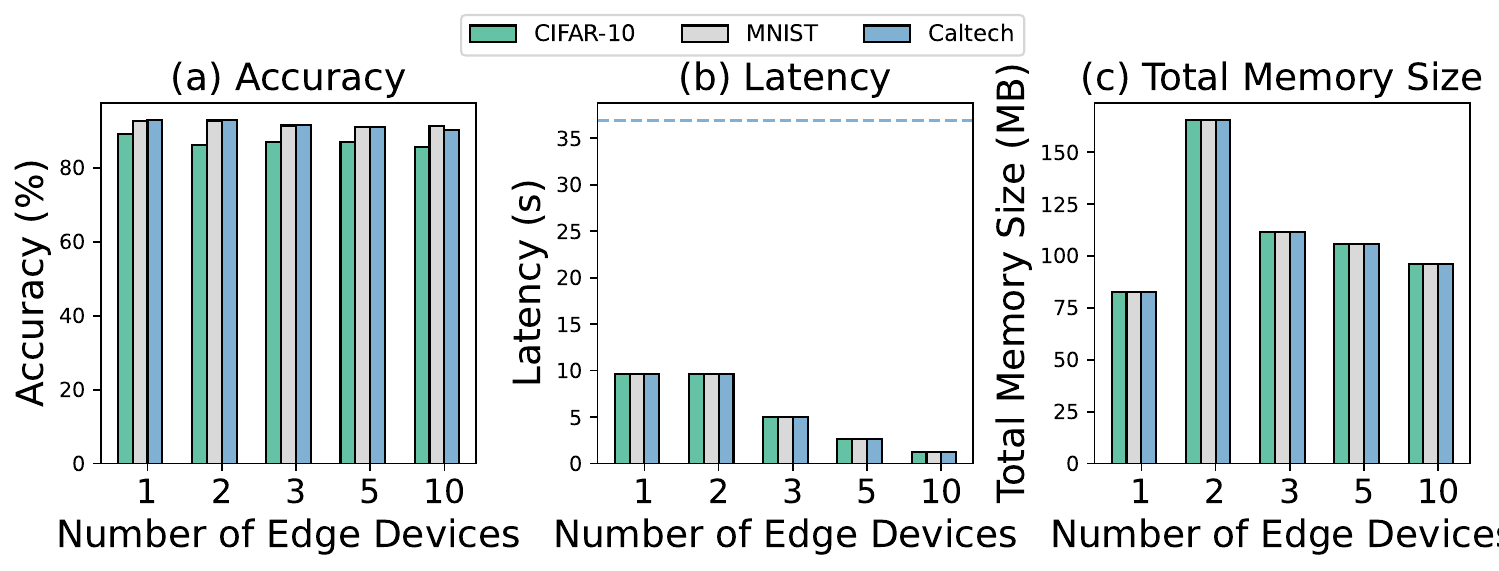}
\caption{Performance metrics of Split ViT-Base models on CIFAR-10, MNIST, Caltech dataset. Note that (a) shows the accuracy results; (b) shows the latency results, the dotted lines represent the latency of the original ViT-Base model, and (c) shows the total memory sizes for all the sub-models. All the experiment results are collected on Raspberry Pi-4B.}
\label{fig:vitcv}
\end{figure}

\begin{figure}[t!]
\centering
  \includegraphics[width=\linewidth]{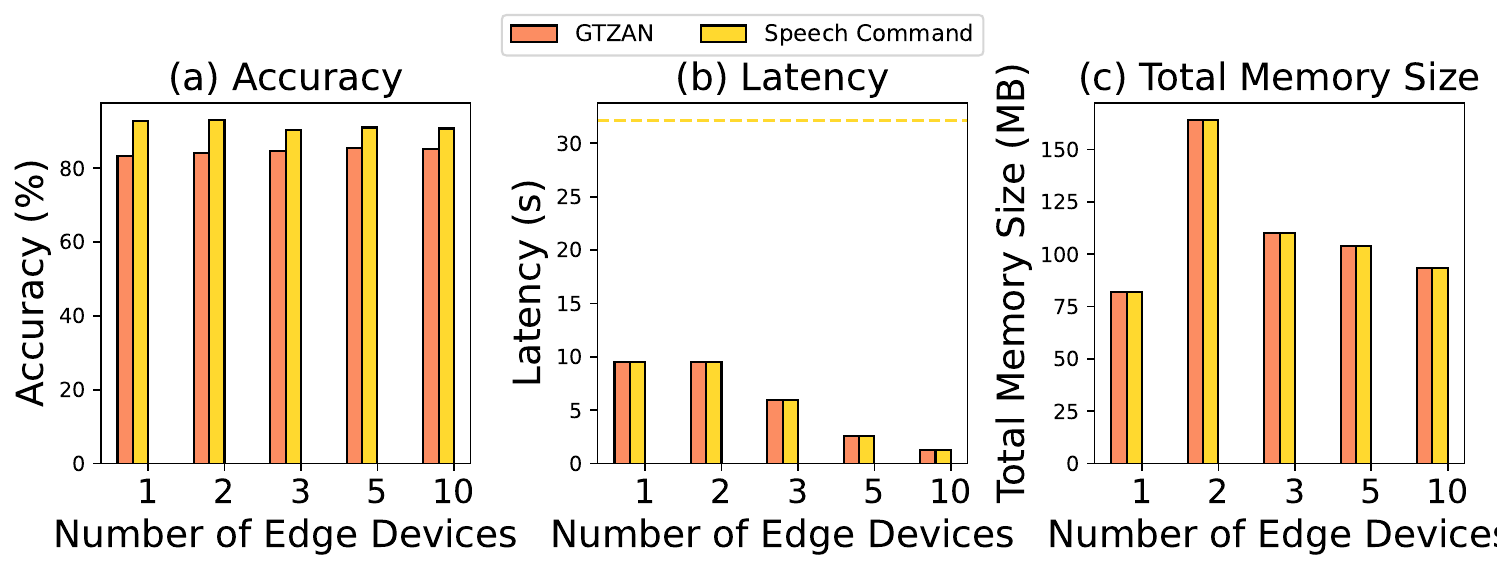}
\caption{Performance metrics of Split ViT-Base models on GTZAN and Speech Command dataset. }
\label{fig:vitaudio}
\end{figure}


\subsection{Experiments on Computer Vision Datasets}
\label{sec:cvexp}

We evaluate our approach using CIFAR-10, MNIST, and Caltech image datasets. The original model size is 327.38 MB. Fig.~\ref{fig:vitcv} shows the accuracy, inference latency, and memory usage of the ViT-Base model under the \edvit framework, with 1 to 10 edge devices. With only one edge device, we apply model compression by pruning \vit without decomposition. All experiments are conducted with a total memory budget of 180MB across devices, ensuring fair comparisons.

The results demonstrate that as the number of edge devices increases, the accuracy remains largely consistent and yields strong performance. For CIFAR-10, accuracies are consistently above 85\%; for MNIST, they are above 91\%; and for Caltech, they exceed 90\%. In most cases, the variance in final fusion prediction accuracy is less than one percentage point. The inclusion of more sub-models illustrates the feasibility of deploying larger-scale models without significant accuracy loss. As the number of edge devices increases, the inference latency decreases, as each sub-model is responsible for fewer classes and contains fewer parameters. Notably, the latency for the original model is 36.94 seconds on the CIFAR-10 dataset, which is 28.9 times the smallest latency (1.28s) and 3.84 times the highest latency (9.63s). 
Our \edvit could make multiple edge devices work collaboratively to maintain accuracy while lowering the storage burden and inference time as the number of edge devices increases. The results for other datasets show a similar trend as the model structures are the same.

In terms of total memory usage, \edvit provides effective splitting and assignment strategies. 
Note that as the number of retained heads increases, the memory size grows quadratically. For one edge device, retaining more heads could exceed the budget. However, in a two-device setting, each sub-model retains a similar number of heads, ensuring the total memory usage remains within the budget. This explains the spike in total memory sizes with two edge devices, as shown in Fig.~\ref{fig:vitcv}.
As the number of edge devices increases from 3 to 10, the memory size of each sub-model decreases, reducing computation overhead and demonstrating that many complex model designs and computational operations are redundant for problem-solving. In the 10-edge device setting, the model size on the CIFAR-10 dataset is reduced to just  9.60MB, achieving a size reduction of up to 34.1 times,  compared to the original model from \edvit.

\begin{figure}[t!]
\centering
  \includegraphics[width=\linewidth]{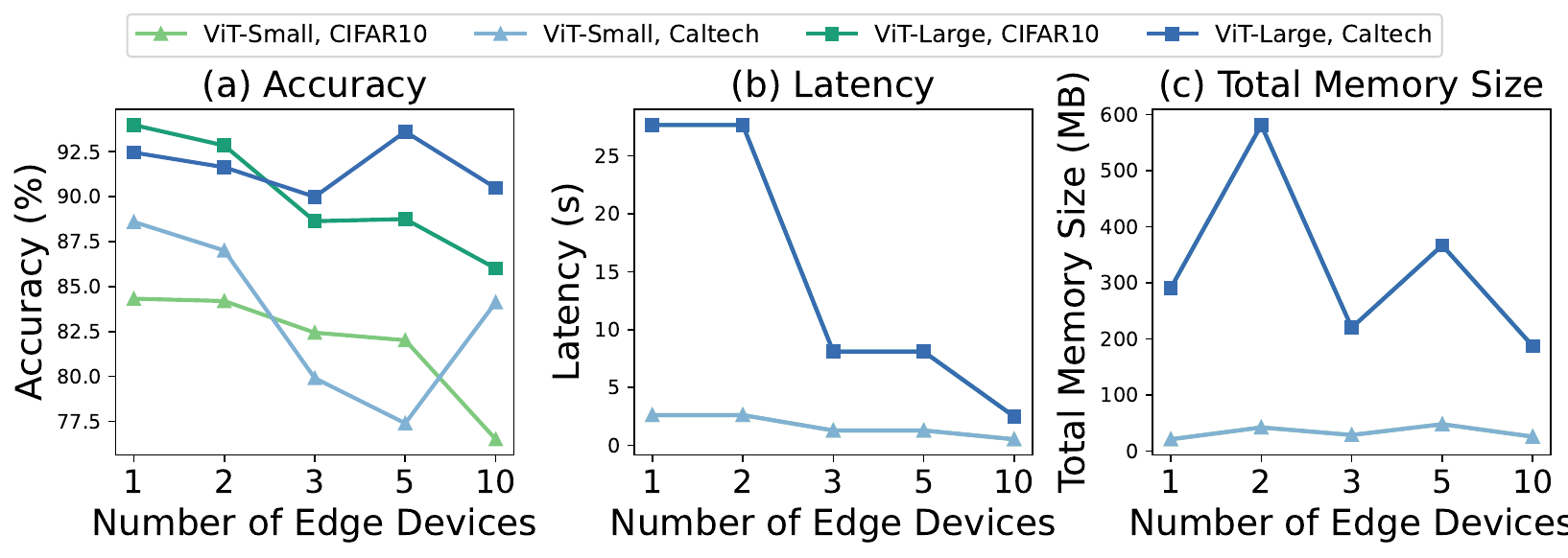}
\caption{Performance metrics of Split ViT-Small and ViT-Large models on CIFAR-10, Caltech dataset.}
\label{fig:vitcvs}
\end{figure}

\subsection{Experiments on Audio Recognition Datasets}
\label{sec:audioexp}

We use the GTZAN and Speech Command audio datasets to evaluate the performance of our framework. The original model sizes of \vit for GTZAN and Speech Command is 325.88MB. Fig.~\ref{fig:vitaudio} presents the accuracy, inference latency, and total memory size of the ViT-Base model as implemented by the \edvit framework, similar to the experiments with the computer vision datasets. We still set the memory budget to 180MB.

The results show that as the number of edge devices increases,  \edvit is able to maintain the accuracy, delivering robust performance. For GTZAN, accuracies are consistently above 84\%, and for the Speech Command dataset, accuracies are above 90\%. Similar to the results on the computer vision datasets, the inference latency decreases as the number of edge devices increases. Notably, the latency for the original model is 32.16 seconds on the GTZAN dataset, which is 25.13 times the smallest latency (1.28s) and 3.37 times the highest latency (9.55s). This substantial latency reduction trend is consistent across both datasets. Regarding total memory size, all configurations remain within the set limits. As the number of edge devices increases, the memory size of each sub-model decreases, and the computation overhead is similarly reduced. In the 10-edge devices setting, for each model in the GTZAN dataset, the size is reduced to only 9.35MB, achieving a reduction of up to 34.85 times compared to the original model. Similar results are also observed in the Speech Command dataset from \edvit. 

\begin{table}[t!]
\centering
\caption{The FLOPs for sub-models on different datasets when using ViT-Base.} 
\begin{tabular}{cccccc}
\hline
\multirow{2}{*}{\textbf{Dataset}} & \multicolumn{5}{c}{The Number of Edge Devices} \\
\cmidrule(r){2-6} 
& Original& 2 &3&5&10 \\ 

\hline
CIFAR-10 & 16.86G & 4.25G &1.90G &1.08G &0.48G \\
GTZAN & 16.79G & 4.20G & 1.88G& 1.059G& 0.46G \\
\hline
\end{tabular}

\label{tab:flops}
\end{table}

\subsection{Overhead of Computation and Communication}

We use FLOPs to simulate the energy consumption on the edge devices. Table~\ref{tab:flops} presents the FLOPs for each edge device on CIFAR (computer vision/video) and GTZAN (audio recognition) datasets with different numbers of edge devices using the ViT-Base model. The FLOPs for the original model on the CIFAR-10 and GTZAN datasets are 16.86G and 16.79G, respectively. As the number of edge devices increases, the FLOPs decrease across all datasets, and the experimental results are consistent with the parameter counts. These findings demonstrate that \edvit significantly reduces computation overhead and saves energy for the edge devices. 

When the number of edge devices increases from 1 to 10 using ViT-Base across all the datasets, the size of features for communication on each sub-model decreases from 1536 bytes to 512 bytes. Compared with the original image size (150528 bytes), our method could greatly reduce the communication overhead to 294 times. The maximal communication time for one edge device is 5.86ms, which is acceptable in the practical situation. The results also show that the inferences on sub-models and the fusion model take up most of the latency (order of seconds) in Section~\ref{sec:cvexp} and Section~\ref{sec:audioexp}.

\subsection{Experiments on Different Model Structures}

We also select two complex datasets (e.g, CIFAR-10 and Caltech) to test different \vit structures for low-power video analytics tasks. The original model sizes of ViT-Small and ViT-Large are 82.71MB, 1,157MB, respectively. Fig.~\ref{fig:vitcvs} presents the accuracy, inference latency, and total memory size of the ViT-Small and ViT-Large models as implemented by the \edvit framework, similar to Fig.~\ref{fig:vitcv}. We increase the total memory size limit for ViT-Large to 600MB and decrease the limit for ViT-Small to 50MB.

The results show that as the number of devices increases, the accuracy remains relatively consistent, again showing robust performance. For ViT-Small, the accuracy is over 76.5\% on the CIFAR-10 dataset and over 77.39\% on Caltech across all settings; for ViT-Large, the accuracy is over 86\% on the CIFAR-10 dataset and over 90.48\% on Caltech in all settings. In most cases, the accuracy fluctuation for the final fusion prediction remains within a variance of less than one percentage point. The accuracy for ViT-Small is lower than that of ViT-Base, while ViT-Large achieves higher accuracy than ViT-Base, corresponding to the difference in parameter counts. Generally, the more parameters, the better the accuracy. As the number of edge devices increases, the latency decreases for both settings, similar to ViT-Base. The latency for ViT-Small is lower than that of ViT-Base, as ViT-Small requires less computational power, while the latency for ViT-Large is higher due to its larger size. In terms of memory size, in the 10-edge device setting, for each model on the CIFAR-10 dataset, the size for ViT-Small is 2.58MB, achieving a reduction of up to 32.06 times compared to the original model. Similarly, for ViT-Large, the size is 18.73MB, which also achieves a 61.77-fold reduction compared to the original model size. 

Note that for the ViT-Small on the CIFAR-10 and CalTech, the input size and the output size are the same; thus, their latency and total memory size on the edge devices are also the same. Similar results are observed across both datasets for ViT-Small and ViT-Large.

\begin{table}[t]
\caption{The accuracy results of splitting CNN and SNN versus \edvit with ViT-Base on CIFAR-10 dataset.}
\label{tab:snncnn}
\centering
\resizebox{\linewidth}{!}{
\begin{tabular}{cccccc}
\toprule
\multirow{2}{*}{\textbf{Methods}} & \multicolumn{5}{c}{The Number of Edge Devices} \\
\cmidrule(r){2-6} 
& 1& 2 &3&5&10 \\ \midrule
Split-CNN & 85.05\stdmode{$\pm$0.32} & 85.11\stdmode{$\pm$0.30} & 85.17\stdmode{$\pm$0.27} & 85.33\stdmode{$\pm$0.14} & 85.31\stdmode{$\pm$0.29}\\ 
Split-SNN & 83.56\stdmode{$\pm$0.01} & 82.45\stdmode{$\pm$0.12} & 83.01\stdmode{$\pm$0.79} & 83.06\stdmode{$\pm$0.50} & 82.29\stdmode{$\pm$0.32}\\ 
\rowcolor[gray]{0.9} 
\edvit & 89.11\stdmode{$\pm$0.40} & 86.18\stdmode{$\pm$0.18} & 86.97\stdmode{$\pm$0.21} & 86.94\stdmode{$\pm$0.19} & 85.59\stdmode{$\pm$0.33}\\ 
\bottomrule
\end{tabular}}
\end{table}

\subsection{Comparison with Baseline Methods: Split-CNN and Split-SNN}
\label{sec:snncnn}
\vit achieves better accuracy compared to traditional CNN and SNN models. However, the performance of these models on edge devices has not been directly compared before. Nnfacet~\cite{chen2023nnfacet} proposes a method to split CNNs across multiple edge devices, employing a filter pruning technique~\cite{Hu2016NetworkTA}, which differs from our approach. EC-SNN~\cite{Yu2024ECSNNSD} utilize the convolutional spiking neural network (CSNN)~\cite{Deng2021OptimalCO} to transform CNNs into SNNs, using a similar strategy. Both methods focus on VGGNet~\cite{simonyan2014very} backbone networks and are channel-wise methods. In our experiments, the baseline model for these methods is VGGNet-16 in their papers, which also has a memory size similar to ViT-Base and achieves the best original results for comparison. We follow the hyper-parameters in their papers to conduct the experiments.

\begin{figure}[t!]
\centering
  \includegraphics[width=\linewidth]{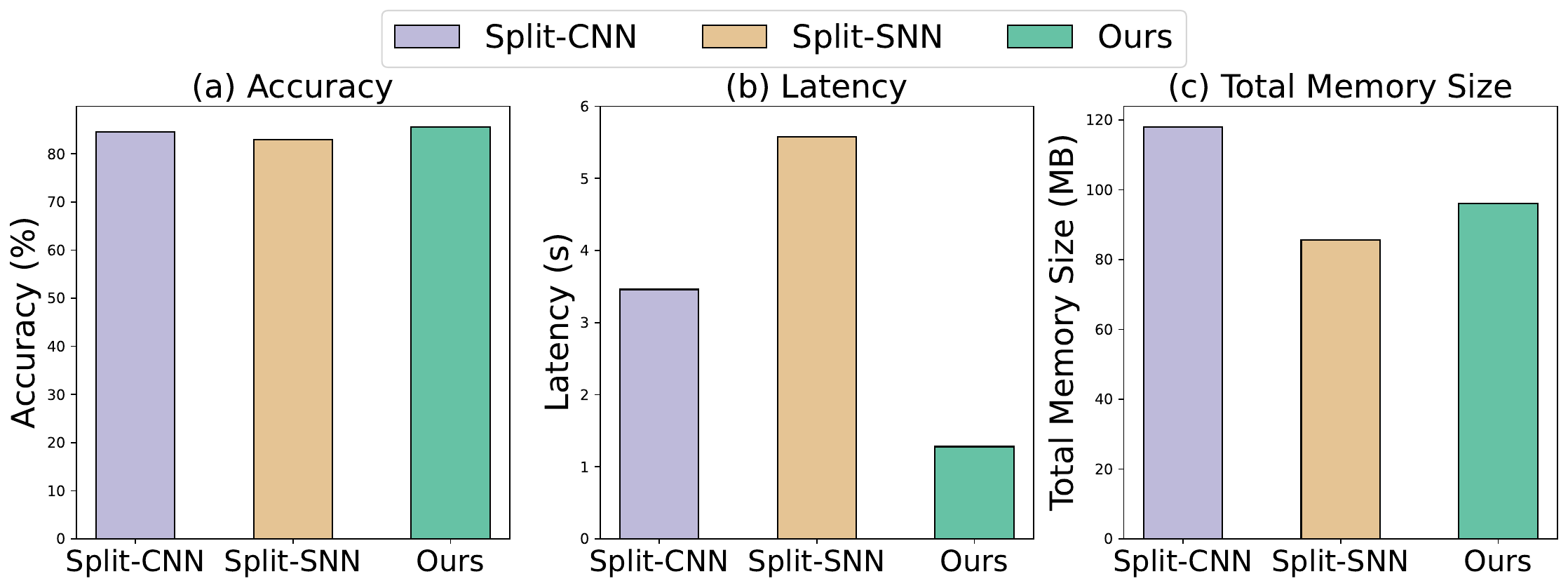}
\caption{Performance of splitting method with CNN, SNN, and ViT-Base models on CIFAR-10 dataset with 10 edge devices. } 
\label{fig:snncnn}
\end{figure}

The accuracy results on the CIFAR-10 dataset are presented in Table~\ref{tab:snncnn}. Based on the results, we observe that CNN outperforms SNN, while our \edvit method for ViT-Base yields better accuracy than both CNN and SNN approaches. 
The original accuracies of ViT-Base, CNN, and SNN are 98.12\%, 93.64\%, and 93.56\%, respectively. Their average accuracy losses are 11.16\%, 8.5\% and 10.15\% across various device numbers. Due to the inherently high accuracy of \vit, we admit that its accuracy drop is slightly higher than CNN and SNN. This is precisely why we leverage \vit: to employ its exceptional performance. Our method maintains a comparable accuracy drop with a 28.9× size reduction, achieving up to 4.06\% and 5.55\% higher accuracy than state-of-the-art CNN-based and SNN-based methods shown in Table~\ref{tab:snncnn}.

In addition to the accuracy results, we also compare inference latency and total memory size of the three methods when the number of edge devices is 10. These results are shown in Fig.~\ref{fig:snncnn}. Based on the results, our \edvit method achieves the best accuracy compared to SNN and CNN, while its inference latency is much lower than SNN (4.36 times) and CNN (2.70 times). Furthermore, the total memory size of \edvit is significantly lower than CNN and is comparable to SNN, since SNN is known for its small model size. This experiment demonstrates that deploying \vit onto edge devices can meet latency and memory constraints while delivering superior accuracy results.

\begin{table}[t]
\caption{The impact of retraining for CIFAR-10 dataset on ViT-Base of \edvit.}
\label{tab:retrain}
\centering
\resizebox{\linewidth}{!}{
\begin{tabular}{lccccc}
\toprule
\multirow{2}{*}{\textbf{Methods}} & \multicolumn{5}{c}{The Number of Edge Devices} \\
\cmidrule(r){2-6} 
& 1& 2 &3&5&10 \\
\midrule
\edvit & 89.11\stdmode{$\pm$0.02}& 86.18\stdmode{$\pm$0.34} & 86.97\stdmode{$\pm$0.70} & 86.94\stdmode{$\pm$0.10} & 85.59\stdmode{$\pm$0.24}\\
(w/o) retrain &88.25\stdmode{$\pm$0.30} &86.00\stdmode{$\pm$0.45} &86.08\stdmode{$\pm$0.03} & 85.33\stdmode{$\pm$0.11}	& 84.20\stdmode{$\pm$0.14}\\
(w/)  entire retrain &89.11\stdmode{$\pm$0.30}	&92.33\stdmode{$\pm$0.11}	&91.14\stdmode{$\pm$0.45}	&89.97\stdmode{$\pm$0.62}	&90.26\stdmode{$\pm$0.07}\\
\bottomrule
\end{tabular}}
\end{table}

\subsection{Experiments on Effects for Retraining}

As we quantify model accuracy, we perform an ablation study to assess the impact of retraining. The results are shown in Table~\ref{tab:retrain}. The first line shows the results of the original \edvit. The second line shows the results from averaging the softmax output of sub-models without the fusion MLP. The third line shows the results based on the retraining of the overall models (sub-models and MLP together) for the fusion stage. When using only one device, the result is the same as the original \edvit as the training process remains unchanged in this scenario. Different from the work on splitting SNN and CNN, which are based on channel-wise methods and only get about 0.1\% improvement in performance when retaining the overall models, our method is shown to have a great potential to improve performance (up to 6.15\%). However, in the practical setting, it may be hard to retrain the sub-models with the fusion MLP.


\section{Conclusion}
\label{sec:conclusion}
In this study, we are the first to propose a novel framework aimed at deploying \vit on edge devices, which combines model-partitioning and pruning. The formulation and resolution of the problem offer a viable solution, \edvit, which decomposes the \vit model into smaller sub-models and leverages the state-of-the-art pruning method to streamline the complex network architecture. \edvit not only preserves the essential structure of the original model but also enables more efficient inference, maintaining high system accuracy within the memory and energy constraints of edge devices.  Extensive experiments and implementations have been conducted on five datasets, three ViT architectures, and two baseline methods, using three evaluation metrics of accuracy, inference latency, and total memory size. The results demonstrate that \edvit significantly reduces overall energy consumption and inference latency on edge devices while maintaining high accuracy. Our \edvit shows great potential for deployment on edge devices and for future integration with other horizontal methods to achieve better performance.


\begin{thebibliography}{10}
\providecommand{\url}[1]{#1}
\csname url@samestyle\endcsname
\providecommand{\newblock}{\relax}
\providecommand{\bibinfo}[2]{#2}
\providecommand{\BIBentrySTDinterwordspacing}{\spaceskip=0pt\relax}
\providecommand{\BIBentryALTinterwordstretchfactor}{4}
\providecommand{\BIBentryALTinterwordspacing}{\spaceskip=\fontdimen2\font plus
\BIBentryALTinterwordstretchfactor\fontdimen3\font minus \fontdimen4\font\relax}
\providecommand{\BIBforeignlanguage}[2]{{%
\expandafter\ifx\csname l@#1\endcsname\relax
\typeout{** WARNING: IEEEtran.bst: No hyphenation pattern has been}%
\typeout{** loaded for the language `#1'. Using the pattern for}%
\typeout{** the default language instead.}%
\else
\language=\csname l@#1\endcsname
\fi
#2}}
\providecommand{\BIBdecl}{\relax}
\BIBdecl

\bibitem{chen2021split}
J.~Chen, D.~Van~Le, R.~Tan, and D.~Ho, ``Split convolutional neural networks for distributed inference on concurrent iot sensors,'' in \emph{2021 IEEE 27th International Conference on Parallel and Distributed Systems (ICPADS)}.\hskip 1em plus 0.5em minus 0.4em\relax IEEE, 2021, pp. 66--73.

\bibitem{chen2023nnfacet}
------, ``Nnfacet: Splitting neural network for concurrent smart sensors,'' \emph{IEEE Transactions on Mobile Computing}, vol.~23, no.~2, pp. 1627--1640, 2023.

\bibitem{Yu2024ECSNNSD}
\BIBentryALTinterwordspacing
D.~Yu, X.~Du, L.~Jiang, W.~Tong, and S.~Deng, ``Ec-snn: Splitting deep spiking neural networks for edge devices,'' \emph{Proceedings of the Thirty-ThirdInternational Joint Conference on Artificial Intelligence}, 2024. [Online]. Available: \url{https://api.semanticscholar.org/CorpusID:271507864}
\BIBentrySTDinterwordspacing

\bibitem{krizhevsky2012imagenet}
A.~Krizhevsky, I.~Sutskever, and G.~E. Hinton, ``Imagenet classification with deep convolutional neural networks,'' \emph{Advances in neural information processing systems}, vol.~25, 2012.

\bibitem{simonyan2014very}
K.~Simonyan and A.~Zisserman, ``Very deep convolutional networks for large-scale image recognition,'' \emph{arXiv preprint arXiv:1409.1556}, 2014.

\bibitem{He2015DeepRL}
\BIBentryALTinterwordspacing
K.~He, X.~Zhang, S.~Ren, and J.~Sun, ``Deep residual learning for image recognition,'' \emph{2016 IEEE Conference on Computer Vision and Pattern Recognition (CVPR)}, pp. 770--778, 2015. [Online]. Available: \url{https://api.semanticscholar.org/CorpusID:206594692}
\BIBentrySTDinterwordspacing

\bibitem{Deng2021OptimalCO}
\BIBentryALTinterwordspacing
S.-W. Deng and S.~Gu, ``Optimal conversion of conventional artificial neural networks to spiking neural networks,'' \emph{ArXiv}, vol. abs/2103.00476, 2021. [Online]. Available: \url{https://api.semanticscholar.org/CorpusID:232075977}
\BIBentrySTDinterwordspacing

\bibitem{vaswani2017attention}
A.~Vaswani, ``Attention is all you need,'' \emph{Advances in Neural Information Processing Systems}, 2017.

\bibitem{alexey2020image}
D.~Alexey, ``An image is worth 16x16 words: Transformers for image recognition at scale,'' \emph{arXiv preprint arXiv: 2010.11929}, 2020.

\bibitem{wu2020lite}
Z.~Wu, Z.~Liu, J.~Lin, Y.~Lin, and S.~Han, ``Lite transformer with long-short range attention,'' \emph{arXiv preprint arXiv:2004.11886}, 2020.

\bibitem{touvron2021training}
H.~Touvron, M.~Cord, M.~Douze, F.~Massa, A.~Sablayrolles, and H.~J{\'e}gou, ``Training data-efficient image transformers \& distillation through attention,'' in \emph{International conference on machine learning}.\hskip 1em plus 0.5em minus 0.4em\relax PMLR, 2021, pp. 10\,347--10\,357.

\bibitem{carion2020end}
N.~Carion, F.~Massa, G.~Synnaeve, N.~Usunier, A.~Kirillov, and S.~Zagoruyko, ``End-to-end object detection with transformers,'' in \emph{European conference on computer vision}.\hskip 1em plus 0.5em minus 0.4em\relax Springer, 2020, pp. 213--229.

\bibitem{dai2021up}
Z.~Dai, B.~Cai, Y.~Lin, and J.~Chen, ``Up-detr: Unsupervised pre-training for object detection with transformers,'' in \emph{Proceedings of the IEEE/CVF conference on computer vision and pattern recognition}, 2021, pp. 1601--1610.

\bibitem{yang2021uncertainty}
F.~Yang, Q.~Zhai, X.~Li, R.~Huang, A.~Luo, H.~Cheng, and D.-P. Fan, ``Uncertainty-guided transformer reasoning for camouflaged object detection,'' in \emph{Proceedings of the IEEE/CVF international conference on computer vision}, 2021, pp. 4146--4155.

\bibitem{song2020vr}
Z.~Song, F.~Wu, X.~Liu, J.~Ke, N.~Jing, and X.~Liang, ``Vr-dann: Real-time video recognition via decoder-assisted neural network acceleration,'' in \emph{2020 53rd Annual IEEE/ACM International Symposium on Microarchitecture (MICRO)}.\hskip 1em plus 0.5em minus 0.4em\relax IEEE, 2020, pp. 698--710.

\bibitem{wang2021end}
Y.~Wang, Z.~Xu, X.~Wang, C.~Shen, B.~Cheng, H.~Shen, and H.~Xia, ``End-to-end video instance segmentation with transformers,'' in \emph{Proceedings of the IEEE/CVF conference on computer vision and pattern recognition}, 2021, pp. 8741--8750.

\bibitem{girdhar2019video}
R.~Girdhar, J.~Carreira, C.~Doersch, and A.~Zisserman, ``Video action transformer network,'' in \emph{Proceedings of the IEEE/CVF conference on computer vision and pattern recognition}, 2019, pp. 244--253.

\bibitem{plizzari2021spatial}
C.~Plizzari, M.~Cannici, and M.~Matteucci, ``Spatial temporal transformer network for skeleton-based action recognition,'' in \emph{Pattern recognition. ICPR international workshops and challenges: virtual event, January 10--15, 2021, Proceedings, Part III}.\hskip 1em plus 0.5em minus 0.4em\relax Springer, 2021, pp. 694--701.

\bibitem{gong2022ssast}
Y.~Gong, C.-I. Lai, Y.-A. Chung, and J.~Glass, ``Ssast: Self-supervised audio spectrogram transformer,'' in \emph{Proceedings of the AAAI Conference on Artificial Intelligence}, vol.~36, no.~10, 2022, pp. 10\,699--10\,709.

\bibitem{Pan2021ScalableVT}
\BIBentryALTinterwordspacing
Z.~Pan, B.~Zhuang, J.~Liu, H.~He, and J.~Cai, ``Scalable vision transformers with hierarchical pooling,'' \emph{2021 IEEE/CVF International Conference on Computer Vision (ICCV)}, pp. 367--376, 2021. [Online]. Available: \url{https://api.semanticscholar.org/CorpusID:232290833}
\BIBentrySTDinterwordspacing

\bibitem{Graham2021LeViTAV}
\BIBentryALTinterwordspacing
B.~Graham, A.~El-Nouby, H.~Touvron, P.~Stock, A.~Joulin, H.~J'egou, and M.~Douze, ``Levit: a vision transformer in convnet’s clothing for faster inference,'' \emph{2021 IEEE/CVF International Conference on Computer Vision (ICCV)}, pp. 12\,239--12\,249, 2021. [Online]. Available: \url{https://api.semanticscholar.org/CorpusID:233004577}
\BIBentrySTDinterwordspacing

\bibitem{Zhang2021MultiScaleVL}
\BIBentryALTinterwordspacing
P.~Zhang, X.~Dai, J.~Yang, B.~Xiao, L.~Yuan, L.~Zhang, and J.~Gao, ``Multi-scale vision longformer: A new vision transformer for high-resolution image encoding,'' \emph{2021 IEEE/CVF International Conference on Computer Vision (ICCV)}, pp. 2978--2988, 2021. [Online]. Available: \url{https://api.semanticscholar.org/CorpusID:232404731}
\BIBentrySTDinterwordspacing

\bibitem{Yu2021MetaFormerIA}
\BIBentryALTinterwordspacing
W.~Yu, M.~Luo, P.~Zhou, C.~Si, Y.~Zhou, X.~Wang, J.~Feng, and S.~Yan, ``Metaformer is actually what you need for vision,'' \emph{2022 IEEE/CVF Conference on Computer Vision and Pattern Recognition (CVPR)}, pp. 10\,809--10\,819, 2021. [Online]. Available: \url{https://api.semanticscholar.org/CorpusID:244478080}
\BIBentrySTDinterwordspacing

\bibitem{Yang2021LiteVT}
\BIBentryALTinterwordspacing
C.~Yang, Y.~Wang, J.~Zhang, H.~Zhang, Z.~Wei, Z.~L. Lin, and A.~L. Yuille, ``Lite vision transformer with enhanced self-attention,'' \emph{2022 IEEE/CVF Conference on Computer Vision and Pattern Recognition (CVPR)}, pp. 11\,988--11\,998, 2021. [Online]. Available: \url{https://api.semanticscholar.org/CorpusID:245353696}
\BIBentrySTDinterwordspacing

\bibitem{Tu2022MaxViTMV}
\BIBentryALTinterwordspacing
Z.~Tu, H.~Talebi, H.~Zhang, F.~Yang, P.~Milanfar, A.~C. Bovik, and Y.~Li, ``Maxvit: Multi-axis vision transformer,'' in \emph{European Conference on Computer Vision}, 2022. [Online]. Available: \url{https://api.semanticscholar.org/CorpusID:247939839}
\BIBentrySTDinterwordspacing

\bibitem{Yao2022DualVT}
\BIBentryALTinterwordspacing
T.~Yao, Y.~Li, Y.~Pan, Y.~Wang, X.~Zhang, and T.~Mei, ``Dual vision transformer,'' \emph{IEEE Transactions on Pattern Analysis and Machine Intelligence}, vol.~45, pp. 10\,870--10\,882, 2022. [Online]. Available: \url{https://api.semanticscholar.org/CorpusID:250425982}
\BIBentrySTDinterwordspacing

\bibitem{Pan2023SlideTransformerHV}
\BIBentryALTinterwordspacing
X.~Pan, T.~Ye, Z.~Xia, S.~Song, and G.~Huang, ``Slide-transformer: Hierarchical vision transformer with local self-attention,'' \emph{2023 IEEE/CVF Conference on Computer Vision and Pattern Recognition (CVPR)}, pp. 2082--2091, 2023. [Online]. Available: \url{https://api.semanticscholar.org/CorpusID:258048654}
\BIBentrySTDinterwordspacing

\bibitem{Yuan2021IncorporatingCD}
\BIBentryALTinterwordspacing
K.~Yuan, S.~Guo, Z.~Liu, A.~Zhou, F.~Yu, and W.~Wu, ``Incorporating convolution designs into visual transformers,'' \emph{2021 IEEE/CVF International Conference on Computer Vision (ICCV)}, pp. 559--568, 2021. [Online]. Available: \url{https://api.semanticscholar.org/CorpusID:232307700}
\BIBentrySTDinterwordspacing

\bibitem{Dai2021CoAtNetMC}
\BIBentryALTinterwordspacing
Z.~Dai, H.~Liu, Q.~V. Le, and M.~Tan, ``Coatnet: Marrying convolution and attention for all data sizes,'' \emph{ArXiv}, vol. abs/2106.04803, 2021. [Online]. Available: \url{https://api.semanticscholar.org/CorpusID:235376986}
\BIBentrySTDinterwordspacing

\bibitem{Wang2021PyramidVT}
\BIBentryALTinterwordspacing
W.~Wang, E.~Xie, X.~Li, D.-P. Fan, K.~Song, D.~Liang, T.~Lu, P.~Luo, and L.~Shao, ``Pyramid vision transformer: A versatile backbone for dense prediction without convolutions,'' \emph{2021 IEEE/CVF International Conference on Computer Vision (ICCV)}, pp. 548--558, 2021. [Online]. Available: \url{https://api.semanticscholar.org/CorpusID:232035922}
\BIBentrySTDinterwordspacing

\bibitem{Wang2021PVTVI}
\BIBentryALTinterwordspacing
------, ``Pvt v2: Improved baselines with pyramid vision transformer,'' \emph{Computational Visual Media}, vol.~8, pp. 415 -- 424, 2021. [Online]. Available: \url{https://api.semanticscholar.org/CorpusID:235652212}
\BIBentrySTDinterwordspacing

\bibitem{zhu2021vision}
M.~Zhu, Y.~Tang, and K.~Han, ``Vision transformer pruning,'' \emph{arXiv preprint arXiv:2104.08500}, 2021.

\bibitem{xia2022structured}
M.~Xia, Z.~Zhong, and D.~Chen, ``Structured pruning learns compact and accurate models,'' \emph{arXiv preprint arXiv:2204.00408}, 2022.

\bibitem{liang2022not}
Y.~Liang, C.~Ge, Z.~Tong, Y.~Song, J.~Wang, and P.~Xie, ``Not all patches are what you need: Expediting vision transformers via token reorganizations,'' \emph{arXiv preprint arXiv:2202.07800}, 2022.

\bibitem{Liu2023RevisitingTP}
\BIBentryALTinterwordspacing
Y.~Liu, M.~Gehrig, N.~Messikommer, M.~Cannici, and D.~Scaramuzza, ``Revisiting token pruning for object detection and instance segmentation,'' \emph{2024 IEEE/CVF Winter Conference on Applications of Computer Vision (WACV)}, pp. 2646--2656, 2023. [Online]. Available: \url{https://api.semanticscholar.org/CorpusID:259138783}
\BIBentrySTDinterwordspacing

\bibitem{tang2023dynamic}
Q.~Tang, B.~Zhang, J.~Liu, F.~Liu, and Y.~Liu, ``Dynamic token pruning in plain vision transformers for semantic segmentation,'' in \emph{Proceedings of the IEEE/CVF International Conference on Computer Vision}, 2023, pp. 777--786.

\bibitem{Zheng2022SAViTSV}
\BIBentryALTinterwordspacing
C.~Zheng, Z.~Li, K.~Zhang, Z.~Yang, W.~Tan, J.~Xiao, Y.~Ren, and S.~Pu, ``Savit: Structure-aware vision transformer pruning via collaborative optimization,'' in \emph{Neural Information Processing Systems}, 2022. [Online]. Available: \url{https://api.semanticscholar.org/CorpusID:258509611}
\BIBentrySTDinterwordspacing

\bibitem{song2022cp}
Z.~Song, Y.~Xu, Z.~He, L.~Jiang, N.~Jing, and X.~Liang, ``Cp-vit: Cascade vision transformer pruning via progressive sparsity prediction,'' \emph{arXiv preprint arXiv:2203.04570}, 2022.

\bibitem{xu2022evo}
Y.~Xu, Z.~Zhang, M.~Zhang, K.~Sheng, K.~Li, W.~Dong, L.~Zhang, C.~Xu, and X.~Sun, ``Evo-vit: Slow-fast token evolution for dynamic vision transformer,'' in \emph{Proceedings of the AAAI Conference on Artificial Intelligence}, vol.~36, no.~3, 2022, pp. 2964--2972.

\bibitem{venkataramanan2023skip}
S.~Venkataramanan, A.~Ghodrati, Y.~M. Asano, F.~Porikli, and A.~Habibian, ``Skip-attention: Improving vision transformers by paying less attention,'' \emph{arXiv preprint arXiv:2301.02240}, 2023.

\bibitem{yu2023x}
L.~Yu and W.~Xiang, ``X-pruner: explainable pruning for vision transformers,'' in \emph{Proceedings of the IEEE/CVF conference on computer vision and pattern recognition}, 2023, pp. 24\,355--24\,363.

\bibitem{yu2023unified}
H.~Yu and J.~Wu, ``A unified pruning framework for vision transformers,'' \emph{Science China Information Sciences}, vol.~66, no.~7, p. 179101, 2023.

\bibitem{li2024lors}
J.~Li, Q.~Nie, W.~Fu, Y.~Lin, G.~Tao, Y.~Liu, and C.~Wang, ``Lors: Low-rank residual structure for parameter-efficient network stacking,'' in \emph{Proceedings of the IEEE/CVF Conference on Computer Vision and Pattern Recognition}, 2024, pp. 15\,866--15\,876.

\bibitem{wadekar2022mobilevitv3}
S.~N. Wadekar and A.~Chaurasia, ``Mobilevitv3: Mobile-friendly vision transformer with simple and effective fusion of local, global and input features,'' \emph{arXiv preprint arXiv:2209.15159}, 2022.

\bibitem{pan2022edgevits}
J.~Pan, A.~Bulat, F.~Tan, X.~Zhu, L.~Dudziak, H.~Li, G.~Tzimiropoulos, and B.~Martinez, ``Edgevits: Competing light-weight cnns on mobile devices with vision transformers,'' in \emph{European Conference on Computer Vision}.\hskip 1em plus 0.5em minus 0.4em\relax Springer, 2022, pp. 294--311.

\bibitem{xu2023devit}
G.~Xu, Z.~Hao, Y.~Luo, H.~Hu, J.~An, and S.~Mao, ``Devit: Decomposing vision transformers for collaborative inference in edge devices,'' \emph{IEEE Transactions on Mobile Computing}, 2023.

\bibitem{Oh2022DifferentiallyPC}
\BIBentryALTinterwordspacing
S.~Oh, J.~Park, S.~Baek, H.~Nam, P.~Vepakomma, R.~Raskar, M.~Bennis, and S.-L. Kim, ``Differentially private cutmix for split learning with vision transformer,'' \emph{ArXiv}, vol. abs/2210.15986, 2022. [Online]. Available: \url{https://api.semanticscholar.org/CorpusID:253224400}
\BIBentrySTDinterwordspacing

\bibitem{Almalik2023FeSViBSFS}
\BIBentryALTinterwordspacing
F.~Almalik, N.~Alkhunaizi, I.~Almakky, and K.~Nandakumar, ``Fesvibs: Federated split learning of vision transformer with block sampling,'' in \emph{International Conference on Medical Image Computing and Computer-Assisted Intervention}, 2023. [Online]. Available: \url{https://api.semanticscholar.org/CorpusID:259251916}
\BIBentrySTDinterwordspacing

\bibitem{Oh2024PrivacyPreservingSL}
\BIBentryALTinterwordspacing
S.~Oh, S.~Baek, J.~Park, H.~Nam, P.~Vepakomma, R.~Raskar, M.~Bennis, and S.-L. Kim, ``Privacy-preserving split learning with vision transformers using patch-wise random and noisy cutmix,'' \emph{ArXiv}, vol. abs/2408.01040, 2024. [Online]. Available: \url{https://api.semanticscholar.org/CorpusID:271693565}
\BIBentrySTDinterwordspacing

\bibitem{Su2022AdaptiveST}
\BIBentryALTinterwordspacing
Z.~Su, H.~Zhang, J.~Chen, L.~Pang, C.-W. Ngo, and Y.-G. Jiang, ``Adaptive split-fusion transformer,'' \emph{2023 IEEE International Conference on Multimedia and Expo (ICME)}, pp. 1169--1174, 2022. [Online]. Available: \url{https://api.semanticscholar.org/CorpusID:248392009}
\BIBentrySTDinterwordspacing

\bibitem{Kim2017SplitNetLT}
\BIBentryALTinterwordspacing
J.~Kim, Y.~Park, G.~Kim, and S.~J. Hwang, ``Splitnet: Learning to semantically split deep networks for parameter reduction and model parallelization,'' in \emph{International Conference on Machine Learning}, 2017. [Online]. Available: \url{https://api.semanticscholar.org/CorpusID:12078675}
\BIBentrySTDinterwordspacing

\bibitem{Bakhtiarnia2022DynamicSC}
\BIBentryALTinterwordspacing
A.~Bakhtiarnia, N.~Milo, Q.~Zhang, D.~Bajovi, and A.~Iosifidis, ``Dynamic split computing for efficient deep edge intelligence,'' \emph{ICASSP 2023 - 2023 IEEE International Conference on Acoustics, Speech and Signal Processing (ICASSP)}, pp. 1--5, 2022. [Online]. Available: \url{https://api.semanticscholar.org/CorpusID:248986420}
\BIBentrySTDinterwordspacing

\bibitem{Hou2022DistrEdgeSU}
\BIBentryALTinterwordspacing
X.~Hou, Y.~Guan, T.~Han, and N.~Zhang, ``Distredge: Speeding up convolutional neural network inference on distributed edge devices,'' \emph{2022 IEEE International Parallel and Distributed Processing Symposium (IPDPS)}, pp. 1097--1107, 2022. [Online]. Available: \url{https://api.semanticscholar.org/CorpusID:246486210}
\BIBentrySTDinterwordspacing

\bibitem{green_machines}
V.~Weaver, ``Green machines - energy efficient machines,'' \url{https://web.eece.maine.edu/~vweaver/group/green_machines.html}, 2024, accessed: 13-Sep-2024.

\bibitem{Krizhevsky2009LearningML}
\BIBentryALTinterwordspacing
A.~Krizhevsky, ``Learning multiple layers of features from tiny images,'' 2009. [Online]. Available: \url{https://api.semanticscholar.org/CorpusID:18268744}
\BIBentrySTDinterwordspacing

\bibitem{LeCun1998GradientbasedLA}
\BIBentryALTinterwordspacing
Y.~LeCun, L.~Bottou, Y.~Bengio, and P.~Haffner, ``Gradient-based learning applied to document recognition,'' \emph{Proc. IEEE}, vol.~86, pp. 2278--2324, 1998. [Online]. Available: \url{https://api.semanticscholar.org/CorpusID:14542261}
\BIBentrySTDinterwordspacing

\bibitem{griffin2007caltech}
G.~Griffin, A.~Holub, P.~Perona \emph{et~al.}, ``Caltech-256 object category dataset,'' Technical Report 7694, California Institute of Technology Pasadena, Tech. Rep., 2007.

\bibitem{tzanetakis2002musical}
G.~Tzanetakis and P.~Cook, ``Musical genre classification of audio signals,'' \emph{IEEE Transactions on speech and audio processing}, vol.~10, no.~5, pp. 293--302, 2002.

\bibitem{warden2018speech}
P.~Warden, ``Speech commands: A dataset for limited-vocabulary speech recognition,'' \emph{arXiv preprint arXiv:1804.03209}, 2018.

\bibitem{paszke2019pytorch}
A.~Paszke, S.~Gross, F.~Massa, A.~Lerer, J.~Bradbury, G.~Chanan, T.~Killeen, Z.~Lin, N.~Gimelshein, L.~Antiga \emph{et~al.}, ``Pytorch: An imperative style, high-performance deep learning library,'' \emph{Advances in neural information processing systems}, vol.~32, 2019.

\bibitem{diederik2014adam}
P.~K. Diederik, ``Adam: A method for stochastic optimization,'' \emph{(No Title)}, 2014.

\bibitem{deng2009imagenet}
J.~Deng, W.~Dong, R.~Socher, L.-J. Li, K.~Li, and L.~Fei-Fei, ``Imagenet: A large-scale hierarchical image database,'' in \emph{2009 IEEE conference on computer vision and pattern recognition}.\hskip 1em plus 0.5em minus 0.4em\relax Ieee, 2009, pp. 248--255.

\bibitem{gemmeke2017audio}
J.~F. Gemmeke, D.~P. Ellis, D.~Freedman, A.~Jansen, W.~Lawrence, R.~C. Moore, M.~Plakal, and M.~Ritter, ``Audio set: An ontology and human-labeled dataset for audio events,'' in \emph{2017 IEEE international conference on acoustics, speech and signal processing (ICASSP)}.\hskip 1em plus 0.5em minus 0.4em\relax IEEE, 2017, pp. 776--780.

\bibitem{TCFoundation2022}
T.~L. Foundation, ``Tc-show / manipulate traffic control settings,'' \url{https://www.linux.com/tutorials/tc-show-manipulate-traffic-control-settings/}, 2022, [Online; accessed 10-October-2023].

\bibitem{Hu2016NetworkTA}
\BIBentryALTinterwordspacing
H.~Hu, R.~Peng, Y.-W. Tai, and C.-K. Tang, ``Network trimming: A data-driven neuron pruning approach towards efficient deep architectures,'' \emph{ArXiv}, vol. abs/1607.03250, 2016. [Online]. Available: \url{https://api.semanticscholar.org/CorpusID:2493219}
\BIBentrySTDinterwordspacing

\end{thebibliography}


\end{document}